\documentclass[sigconf]{acmart}

\usepackage{multirow}
\usepackage{algpseudocode} 
\usepackage{algorithm}
\usepackage{makecell}
\usepackage{balance}
\usepackage{threeparttable}
\newtheorem{myDef}{Definition}

\AtBeginDocument{%
  }

\copyrightyear{2025}
\acmYear{2025}
\setcopyright{acmlicensed}\acmConference[CIKM '25]{Proceedings of the 34th ACM International Conference on Information and Knowledge Management}{November 10--14, 2025}{Seoul, Republic of Korea}
\acmBooktitle{Proceedings of the 34th ACM International Conference on Information and Knowledge Management (CIKM '25), November 10--14, 2025, Seoul, Republic of Korea}
\acmDOI{10.1145/3746252.3761141}
\acmISBN{979-8-4007-2040-6/2025/11}

\begin{document}
	
	\title{Structure-Attribute Transformations with Markov Chain Boost Graph Domain Adaptation}

	\author{Zhen Liu}
	\authornote{Both authors contributed equally to this research.}
	\email{quake@uestc.edu.cn}
	\orcid{0000-0001-8762-0664}
	\authornote{Corresponding author.}
	\affiliation{
		\department{School of Computer Science and Engineering}
		\institution{University of Electronic Science and Technology of China}
		\city{ChengDu}
		\state{SiChuan}
		\country{China}
	}
	
	\author{Yongtao Zhang}
	\email{202222080730@std.uestc.edu.cn}
	\orcid{0009-0001-3500-8390}
	\authornotemark[1]
	\affiliation{
		\department{School of Computer Science and Engineering}
		\institution{University of Electronic Science and Technology of China}
		\city{ChengDu}
		\state{SiChuan}
		\country{China}
	}
	
	\author{Shaobo Ren}
	\email{202222081131@std.uestc.edu.cn}
	\orcid{0009-0008-3542-6868}
	\affiliation{
		\department{School of Computer Science and Engineering}
		\institution{University of Electronic Science and Technology of China}
		\city{ChengDu}
		\state{SiChuan}
		\country{China}
	}
	
	\author{Yuxin You}
	\email{202321081209@std.uestc.edu.cn}
	\orcid{0009-0000-7517-0045}
	\affiliation{
		\department{School of Computer Science and Engineering}
		\institution{University of Electronic Science and Technology of China}
		\city{ChengDu}
		\state{SiChuan}
		\country{China}
	}

	\renewcommand{\shortauthors}{Zhen Liu, Yongtao Zhang, Shaobo Ren, and Yuxin You}

	\begin{abstract}
		Graph domain adaptation has gained significant attention in label-scarce scenarios across different graph domains. Traditional approaches to graph domain adaptation primarily focus on transforming node attributes over raw graph structures and aligning the distributions of the transformed node features across networks. However, these methods often struggle with the underlying structural heterogeneity between distinct graph domains, which leads to suboptimal distribution alignment. To address this limitation, we propose Structure-Attribute Transformation with Markov Chain (SATMC), a novel framework that sequentially aligns distributions across networks via both graph structure and attribute transformations. To mitigate the negative influence of domain-private information and further enhance the model's generalization, SATMC introduces a private domain information reduction mechanism and an empirical Wasserstein distance. Theoretical proofs suggest that SATMC can achieve a tighter error bound for cross-network node classification compared to existing graph domain adaptation methods. Extensive experiments on nine pairs of publicly available cross-domain datasets show that SATMC outperforms state-of-the-art methods in the cross-network node classification task. The code is available at https://github.com/GiantZhangYT/SATMC.
	\end{abstract}



	
	\maketitle
	
	\section{Introduction}
	
\begin{figure}[ht]
		\centering
		\includegraphics[width=3.5in]{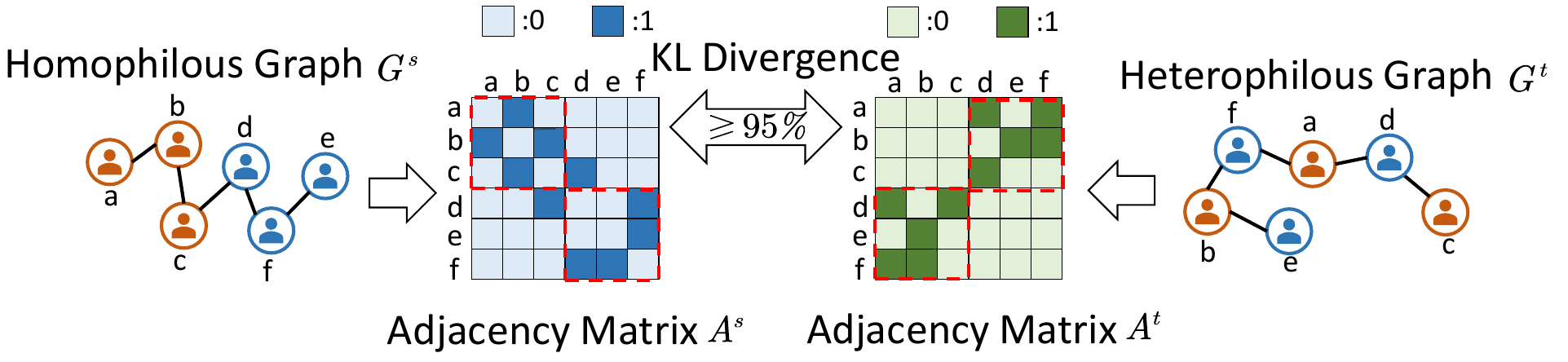}
		\caption{A case for social network analysis in which the source and target graphs have heterogeneous graph topological distributions. }
		\label{motivation}
	\end{figure}

	Traditional Domain Adaptation (DA) methods mainly focused on attributed tabular data, and their goal is to eliminate domain shifts by reducing the discrepancy in attribute feature distributions between the source and target domains \cite{ben2010theory,wilson2020survey,zhou2022domain}. These methods have achieved significant successes in few-shot or zero-shot classification tasks associated with computer vision and natural language processing \cite{peng2018zero,wang2019few,rietzler2019adapt,wang2024dual}. However, for Graph Domain Adaptation (GDA), conventional methods become infeasible, owing to that graph data contains both node features (attribute data) and graph structure features (topological information), which jointly determine the overall properties and functions of the graph. Simply aligning the distributions of node features is not sufficient to fully bridge the gap between the source and target graph domains \cite{pan2018adversarially,zhou2020graph}. As such, GDA needs to eliminate dual domain discrepancies in both node attribute space and structural feature space. From the perspective of the node attribute, graph data from different domains often originates from different generation processes, which may lead to significant differences in the distributions of the node attribute \cite{xu2022graph}. From the structural feature perspective, the topologies of graphs from different domains may have completely different connectivity patterns. For example, in social network analysis, the source domain may be a homophilous graph while the target domain may contain a heterophilous graph. Accordingly, as shown in Fig. \ref{motivation}, the edges of the homophilous graph are mainly distributed in the diagonal area of the adjacency matrix while the edges of the heterophilous graph are primarily distributed in the off-diagonal area. Such structural discrepancies can significantly undermine the learning effectiveness of cross-network node embeddings, thus limiting the performance of the GDA.
	
	Studies have verified that graph structure plays an important role in graph representation learning as it determines the connection patterns between nodes as well as the propagation path of the global information flow \cite{kipf2016semi,hamilton2017inductive}. However, most existing graph domain adaptation methods focus on node attribute transformations on raw graph structures, such as through adversarial learning \cite{dai2022graph} or embedding alignment techniques \cite{chen2023universal,wu2020unsupervised}, to reduce inter-domain distributional divergence. While these methods can be effective in some cases, they generally overlook the role of structure transformations to align the inter-domain structural distributions. When the graph structures of the source and target domains differ significantly such as the case shown in Fig. \ref{motivation}, merely performing node feature transformations on raw graph structures may not be sufficient to achieve consistent representations between domains. Wu et al. \cite{wu2023non} have reported that the generalization performance of the GDA model can be improved through a sort of simple structure transformation, i.e., Weisfeiler-Lehman subtree transformation, when the source and target domains are not independent and identically distributed (i.i.d.). In the task of domain adaptation of molecular graphs, differences in molecular structure directly affect the prediction accuracy of chemical properties \cite{jin2020hierarchical}. Therefore, in GDA, effectively combining graph structure transformation and node feature alignment can better reduce structural and attribute inconsistencies of cross-domain graph data.  
	
	With the aforementioned motivation, we propose a novel framework for GDA based on structure-attribute transformations with a Markov chain. We first utilize graph diffusion to conduct the graph structure transformation. Under the condition that the graph structure distributions are already intended to be consistent, we further perform the graph attribute transformation to promote the node attribute distributions to reach consistency.
	
	Our contributions can be summarized as follows:
	\begin{itemize}
		\item Unlike the traditional GDA methods, SATMC introduces ‌Markov chain boosting‌, which reformulates the graph domain adaptation as a dynamic, multi-step stochastic process and guarantees a tighter error bound for the cross-network node classification.
		
		\item Regarding the common issue that private domain information negatively affects domain distribution alignment \cite{zhang2021adversarial}, we propose a private domain information reduction mechanism based on mutual information minimization.

		\item We propose the empirical Wasserstein-1 distance between the distributions of the source graph and target graph, which corresponds to the complex discrepancies between the graph domains.
		
		\item Extensive experiments for the cross-network node classification task on real-world datasets validated that our model outperforms the state-of-the-art GDA methods.
		
	\end{itemize}

	\section{Related Work}
	\subsection{Domain Adaptation in Graphs}

	Early studies proposed graph-specific Maximum Mean Discrepancy (MMD) adaptations simply from a graph embedding perspective \cite{chen2019graph,li2022structure}. Compared to traditional DA techniques, graph domain adaptation studies are inclined to leverage graph properties from diverse perspectives \cite{shi2025domain} including coarsening approach \cite{li2022domain}, Spectral Augmentation (SA-GDA) for category-level feature alignment \cite{pang2023sa}, category-level domain adaptive embedding \cite{shi2023node}, and Weisfeiler-Lehman test-based cross-network knowledge transferability \cite{wu2023non}. Recently, integrating adversarial learning with graph embedding has gained attention. Models like ACDNE \cite{shen2020adversarial}, ASN \cite{zhang2021adversarial},  LHCDA \cite{xue2023label}, MS-CNC\cite{zhang2023multi} and AdaGCN \cite{dai2022graph} focused on how to apply adversarial learning to graph domain adaptation.
	
	Compared with existing studies which mainly address the graph domain distribution alignment via graph attribute transformation, our proposed structural-attribute distribution alignment framework has better generalization ability.
	
	\subsection{Graph Diffusion}

	Recent advances in computer vision \cite{croitoru2023diffusion}, image processing \cite{tsiotsios2013choice}, and physics-inspired learning \cite{belbute2020combining} have sparked interest in diffusion technology for graph learning. This research line includes neural diffusion on graphs \cite{wang2021dissecting} by solving deep continuous graph diffusion equations \cite{thorpe2021grand++,choi2023climate}. \cite{chamberlain2021grand} proposed GRAND, which parameterizes graph diffusion equations using neural networks, and \cite{chamberlain2021beltrami} introduced a new GNN based on Beltrami neural diffusion for non-Euclidean diffusion in joint position and feature space. \cite{choi2023gread} further explored the impact of combining diffusion equations with reaction mechanisms on graph diffusion, better handling graphs with varying homophily ratios.
	
	These explorations have inspired new GNN designs. However, the role of graph diffusion in graph domain adaptation remains unclear, and our work aims to bridge this gap.
	
	\subsection{Methodologies for Cross-Network Node Classification}
	
	For cross-network node classification, there are two main categories: semi-supervised domain adaptation (SSDA) and unsupervised domain adaptation (UDA). To tackle SSDA with limited labeled nodes in the target graph, \cite{xiao2024semi} proposed SemiGCL, which uses graph contrastive learning and minimax entropy training. Compared to SSDA, most works focus on the UDA for a more general purpose. For instance, \cite{cai2024graph} implements UDA by introducing a variational graph autoencoder to disentangle three independent latent variables. \cite{liu2024structure} developed the structurally-enhanced prototype alignment (SEPA) framework to learn domain-invariant representations of non-i.i.d. data. \cite{wu2020unsupervised} proposed the unsupervised domain adaptive graph convolutional network (UDAGCN), which features a dual-graph convolutional network for joint local and global feature aggregation with attention mechanisms.
	
	The prior endeavors seldom relieve the side effects of the private domain information during UDA. Aiming at this issue, we proposed to explicitly isolate private domain information.

	\section{Preliminary}
	
	\textbf{Cross-network node classification:} 
	There are a fully labeled source attribute graph $G^{s} = \left( V^{s},E^{s},X^{s} \right)$ with a node label matrix $Y^{s}$ and an unlabeled target attribute graph $G^{t} = \left( V^{t},E^{t},X^{t} \right)$, which share the same or overlapped label space $\mathbb{C} = \left\{ 1,2, \cdot \cdot \cdot ,c \right\}$ \cite{chen2023universal}. Suppose $G^{s}$ and $G^{t}$ come from two different distributions $\mathbb{P} _{G^{s}}$ and $\mathbb{P} _{G^{t}}$, cross-network node classification aims to build a classifier $h$ to classify unlabeled nodes in the target graph. Since there exist domain shifts between the source graph and the target graph, i.e., $\mathbb{P} _{G^{s}}\ne \mathbb{P} _{G^{t}}$, to make the classifier $h$ effective on the target graph, we need to adopt an appropriate graph domain adaptation strategy to align the two graphs in the domain space. According to Theorem 1 in \cite{wu2023non}, given a graph domain discriminator $\mathcal{D}(\mathbb{P} _{G^s},\mathbb{P} _{G^t})$ to measure the domain shift between the distributions of $\mathbb{P} _{G^s}$ and $\mathbb{P} _{G^t}$, the classification error on the target nodes can be bounded as
	\begin{equation}
		\epsilon_{t}(h) \le \epsilon_{s}(h)+\mathcal{D}(\mathbb{P} _{G^s},\mathbb{P} _{G^t})+\lambda^{*}+R^{*}
		\label{berror}
	\end{equation}
	where $\lambda^{*}=\epsilon_{t}(h^{*}_{s},h^{*}_{t})$ measures the prediction bias of optimal source and target classifiers on the target nodes, and $R^{*}=\epsilon_{s}(h^{*}_{s})+\epsilon_{t}(h^{*}_{t})$ is the Bayes error on the source and target graphs. Given that $\lambda^{*}$ and $R^{*}$ are nearly fixed, the objective of $\min \epsilon_{t}(h) $ lies in the optimization of the first term and the second term on the right side of the inequality (\ref{berror}). We mainly focus on the second term because the first term is closely dependent on the second term.\\
	\textbf{Graph domain adaptation (GDA) in latent space}: 
	It is not convenient to build a domain discriminator directly between the graphs due to their non-Euclidean characteristics. Thanks to the rapid development of graph representation learning technology in recent years \cite{ju2024comprehensive,khoshraftar2024survey}, one can map the graph to a unified latent space and build a graph domain discriminator in the latent space to reduce the graph domain discrepancy. Thus, graph domain adaptation aims to minimize the probability distribution distance between the source graph $G^s$ and the target graph $G^t$ in the latent space, which can be formulated as

	\begin{equation}
		\begin{aligned}
			&\min  \mathcal{D}\left(\mathbb{P}_{s},\mathbb{P}_{t}\right)\\
			&s.t.~~\begin{cases}
				\phi_{s} :G^s \to \hat{X}^{s} \sim \mathbb{P}_{s}\\
				\phi_{t} :G^t \to \hat{X}^{t} \sim \mathbb{P}_{t}
			\end{cases},
		\end{aligned}
		\label{evolution}
	\end{equation}
	where $\phi_{s}$ and $\phi_{t}$ are graph domain projectors to map the source graph and target graph into latent feature matrices $\hat{X}^{s}$ and $\hat{X}^{t}$, assuming that $\hat{X}^{s}$ follows a marginal probability distribution $\mathbb{P}_{s}$ and $\hat{X}^{t}$ follows another marginal distribution $\mathbb{P}_{t}$, i.e., $\hat{X}^{s} \sim \mathbb{P}_{s}$, $\hat{X}^{t} \sim \mathbb{P}_{t}$, and $\mathbb{P}_{s}\ne \mathbb{P}_{t}$. If, after the graph domain adaptation, the source and target graph domains could (approximately) reach the identical distribution, i.e., $\mathbb{P}_{s}=\mathbb{P}_{t}$, then $\hat{X}^{s}$ and $\hat{X}^{t}$ are called \textbf{domain-invariant representations}. As a result, compared to non-domain-invariant representations without GDA, a better semi-supervised classifier $h$ can be built upon the $\hat{X}^{s}\bigcup\hat{X}^{t}$ under the i.i.d. condition.

	\section{Methodology}
	
	Unlike most traditional studies that commonly overlook the domain discrepancy in graph structural distribution, we design three optimization paths for graph domain adaptation by simultaneously considering graph structure alignment and graph attribute alignment.
	
	\subsection{Optimization Paths of GDA}
	
	Since attribute graphs have both structural and attribute features, the inconsistency of graph distributions is supposed to stem from the joint effects of structural feature shift and attribute feature shift, i.e., $\mathbb{P}_{A^s,X^s} \neq \mathbb{P}_{A^t,X^t}$. Thus, we need to conduct structural and attribute transformation simultaneously such that $\mathbb{P}_{\hat{A}^s,\hat{X}^s}= \mathbb{P}_{\hat{A}^t,\hat{X}^t}$, where $\hat{A}^s$,$\hat{A}^t$ are the transformed adjacency matrices and $\hat{X}^s,\hat{X}^t$ are the transformed attribute matrices. As joint distributions satisfy $\mathbb{P}_{\hat{A}^s,\hat{X}^s}=\mathbb{P}_{\hat{X}^s|\hat{A}^s}\mathbb{P}_{\hat{A}^s}$ and $\mathbb{P}_{\hat{A}^t,\hat{X}^t}=\mathbb{P}_{\hat{X}^t|\hat{A}^t}\mathbb{P}_{\hat{A}^t}$, the cross-network graph distribution alignment problem can be converted as structure and attribute transformations in sequence.
	\begin{equation}
		\begin{aligned}
			&\min D_{\text{KL}}(\mathbb{P}_{\hat{X}^{s}|\hat{A}^{s}}\parallel\mathbb{P}_{\hat{X}^{t}|\hat{A}^{t}})\\
			&s.t.~~\begin{cases}
				\hat{A}^{s}=\Upsilon_{s}^{T} A^{s}\Upsilon_{s}\\
				\hat{A}^{t}=\Upsilon_{t}^{T} A^{t}\Upsilon_{t}\\
				\hat{X}^{s}=\text{T\small{F}}(X^{s}\succ \hat{A}^{s})\\
				\hat{X}^{t}=\text{T\small{F}}(X^{t}\succ \hat{A}^{t})
			\end{cases},
		\end{aligned}
		\label{m3}
	\end{equation}
	where $\Upsilon_{s}$ and $\Upsilon_{t}$ are permutation matrices and $D_{\text{KL}}(\cdot)$ is the Kullback-Leiber divergence. Given the transformed adjacency matrices $\hat{A}^{s}$ and $\hat{A}^{t}$, the attribute shifts are further reduced through graph attribute transformation where $\text{T\small{F}}(\cdot)$ is an attribute transformation function. Therefore, for both source and target graphs, the graph transformation can be conducted through a Markov chain $(A,X)\rightarrow (\hat{A},X)\rightarrow (\hat{A},\hat{X})$. Due to the Markovian property, the joint distribution alignment problem can be relaxed to the posterior distribution alignment problem.

	On the other hand, since the transformed node representation matrix always retains the private feature information of a given graph domain, we need to eliminate such information when aligning the domain distributions. Assuming that the graph has two representations that follow the distributions $\mathbb{P}_{\tilde{X}}$ and $\mathbb{P}_{\hat{X}}$, where the former represents the prior private distribution of the graph domain and the latter is the posterior distribution after domain adaptation. We define the Markov-chain $(A,X)\rightarrow (A,\tilde{X})$ to derive the prior graph presentation with only graph attribute transformation. To reduce the private distribution information of $\hat{X}$, we aim to minimize the mutual information between $\hat{X}$ and $\tilde{X}$ by calculating $I(\hat{X}; \tilde{X})=H(\hat{X})-H(\hat{X}|\tilde{X})$. Thus, we have the equivalent optimization functions for the source and target graphs
	
	\begin{equation}
		\begin{aligned}
			\min D_{\text{KL}}(\mathbb{P}_{\hat{X}^s|\tilde{X}^s}\parallel\mathbb{P}_{\hat{X}^s}),\\
			\min D_{\text{KL}}(\mathbb{P}_{\hat{X}^t|\tilde{X}^t}\parallel\mathbb{P}_{\hat{X}^t}).
		\end{aligned}
		\label{m4}
	\end{equation}
	
	Finally, for the cross-network node classification problem, we also need to perform domain alignment in the label space. Assuming that the node label matrix $Y^s$ of the source graph and predicted node label matrix $\hat{Y}^t$ of the target graph have the distributions $\mathbb{P}_{Y^s}$ and $\mathbb{P}_{\hat{Y^t}|\hat{X^t}}$, the goal of label domain alignment aims to 
	\begin{equation}
		\min D_{\text{KL}}(\mathbb{P}_{Y^s} \parallel\mathbb{P}_{\hat{Y^t}|\hat{X^t}}).
		\label{m5}
	\end{equation}
	
	By putting them together, the joint optimal objectives of the GDA are formulated as
	\begin{equation}
		\begin{aligned}
			\min [D_{\text{KL}}(\mathbb{P}_{\hat{X}^{s}|\hat{A}^{s}}\parallel\mathbb{P}_{\hat{X}^{t}|\hat{A}^{t}})
			+D_{\text{KL}}(\mathbb{P}_{\hat{X}^s|\tilde{X}^s}\parallel\mathbb{P}_{\hat{X}^s})\\+D_{\text{KL}}(\mathbb{P}_{\hat{X}^t|\tilde{X}^t}\parallel\mathbb{P}_{\hat{X}^t})+D_{\text{KL}}(\mathbb{P}_{Y^s} \parallel\mathbb{P}_{\hat{Y^t}|\hat{X^t}})].
		\end{aligned}
		\label{m6}
	\end{equation}
	
	In a nutshell, the GDA objective involves three optimization paths.
	\begin{itemize}
		\item {} Markov-chain-based graph transformation for structure-attribute alignment (the first item in Eq. (\ref{m6})).
		\item {} Domain private information reduction based on mutual information minimization (the second and third items in Eq. (\ref{m6})).
		\item {} Label space alignment (the fourth item in Eq. (\ref{m6})).
	\end{itemize}

\begin{figure}
		\center
		\includegraphics[width=1\linewidth]{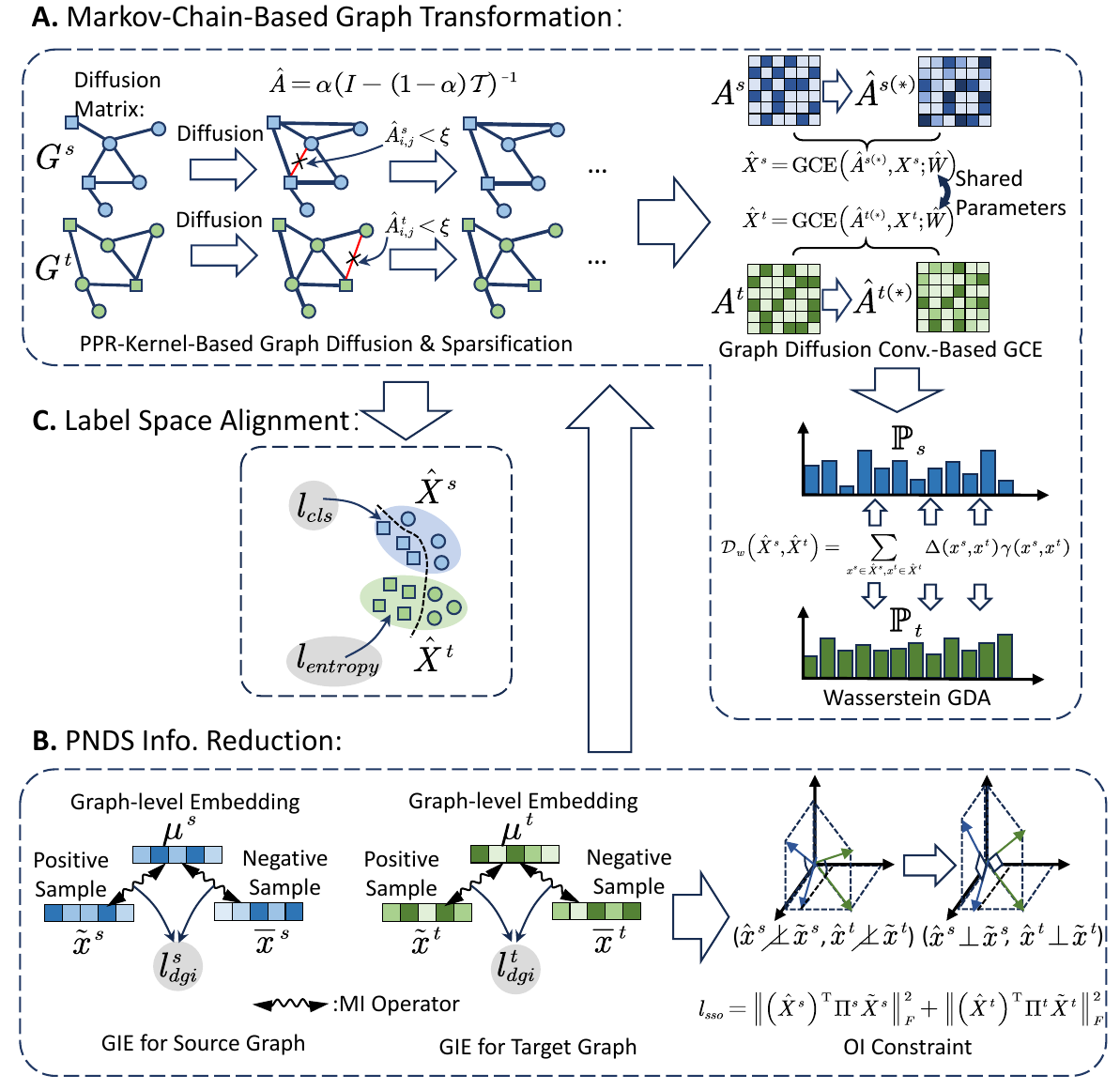}
		\caption{Computational framework of SATMC. (A) Markov-chain-based graph transformation. (B) PNDS info. reduction.(C) Label space alignment.}
		\label{Framework_fig}
	\end{figure}

	Consequently, we propose a novel graph domain adaptation framework, dubbed Structural-Attribute Transformation with Markov-Chain (SATMC), shown in Fig. \ref{Framework_fig}.
	\subsection{Markov-Chain-Based Graph Transformation}
	To realize the first optimization path, we need to perform Markov-chain-based graph transformations. However, in formula (\ref{m3}), finding optimal permutation matrices $\Upsilon^{s}$ and $\Upsilon^{t}$ is an NP-hard problem. Instead, we propose an approximate solution using a cross-network graph consistency encoder (GCE). The GCE aims to capture cross-network domain-invariant (CNDI) information across the source and target graphs.

	Most existing studies trivially use GNNs to extract node features based on the raw graph structures \cite{chen2023universal,dai2022graph,shen2023domain}, which might contain heterogeneous topological information. In contrast, we consider collecting node features on the transformed graph structures that could capture more graph domain-invariant information. According to the Markov chain $(A,X)\rightarrow (\hat{A},X)\rightarrow (\hat{A},\hat{X})$, inspired by the generative graph methods \cite{chamberlain2021beltrami,chamberlain2021grand}, the graph structural-attribute transformations are performed by Neural Graph Diffusion Convolution (NGDC) with Personalized PageRank (PPR) kernel \cite{page1998pagerank}. Thus, the GCE encoders with shared parameter matrix $\hat{W}^{(i)}$ have the forms

	\begin{equation}
		\begin{aligned}		
			&\hat{X}^{s(i+1)} = \text{GCE}\left( \hat{A}^{s(*)},\hat{X}^{s(i)},\hat{W}^{(i)} \right),\\
			&\hat{X}^{t(i+1)} = \text{GCE}\left( \hat{A}^{t(*)},\hat{X}^{t(i)},\hat{W}^{(i)} \right),
		\end{aligned}
		\label{gce}
	\end{equation}
	where $\hat{X}^{s(0)}=X^{s}$ and $\hat{X}^{t(0)}=X^{t}$, respectively. The transformed adjacency matrices $\hat{A}^{s(*)}$ and $\hat{A}^{t(*)}$ are obtained by the processes of graph structural diffusion and sparsification, which can be referred to \cite{gasteiger2019diffusion} for the details. The superscript $i+1$ represents the output of the $i+1$th layer of the GCE. For computational convenience, we simply adopt GCN to conduct graph convolution operations \cite{kipf2016semi}.
	
	Due to the diversity of the distributions and the non-Euclidean nature of the graphs, simply reducing the Euclidean distance between distributions will lead to suboptimal results \cite{you2023graph}. Inspired by the Optimal Transport (OT) theory \cite{villani2021topics,peyre2019computational}, we choose the $p$-th Wasserstein distance as the graph domain discriminator to measure the complex discrepancies between graph domains, which is defined as
	\begin{equation}
		\mathcal{D}_{w}(\hat{X}^{s},\hat{X}^{t}) = \left( \inf_{\gamma \in \Gamma(P, Q)} \int_{\Omega \times \Omega} \Delta(x^s, x^t)^p \, d\gamma(x^s, x^t) \right)^{1/p},
		\label{wass}
	\end{equation}
	where $x^s\in \hat{X}^{s}$, $x^t\in \hat{X}^{t}$, and $\gamma(x^s, x^t)$ is a selected policy for transporting one unit of material from the position $x^s$ to another position $x^t$ with cost $\Delta(x^s, x^t)^p$. $\mathcal{D}_{w}(\hat{X}^{s},\hat{X}^{t})$ represents the cumulated transport cost. Via reducing the $p$-th Wasserstein distance, $\hat{X}^{s}$ and $\hat{X}^{t}$ will more consistent distributions. We call the corresponding graph domain adaptation approach as \textbf{Wasserstein GDA} or \textbf{WGDA}.\\
	\textbf{Empirical graph Wasserstein distance:} It is intractable to directly use the formula (\ref{wass}) to calculate the Wasserstein distance without knowing the actual distributions of the source and target graph domains. To solve this problem, we make an assumption about graph embedding distributions from an empirical perspective. Suppose the source graph embedding and target graph embedding follows multivariate normal distributions, i.e., $\hat{X}^{s}\sim N(\mu^{s},\Sigma^{s})$, $\hat{X}^{t}\sim N(\mu^{t},\Sigma^{t})$, where $\mu^{s}$, $\mu^{t}$ are the mean embeddings and $\Sigma^{s}$, $\Sigma^{t}$ are the covariance matrices. The empirical Wasserstein-1 distance between the source graph and the target graph in the embedding space has a closed-form expression 
	\begin{equation}
		\begin{aligned}
			\mathcal{D}_{w}(\hat{X}^{s},\hat{X}^{t})= &\sum_{x^s\in \hat{X}^{s}, x^t\in \hat{X}^{t}}{\Delta(x^s,x^t)\gamma(x^s,x^t)}\\
			= &\|\mu^{s}-\mu^{t}\|+\mathrm{Tr}(\Sigma^{s})+\mathrm{Tr}(\Sigma^{t})\\
			&-2\mathrm{Tr}[(\sqrt{\Sigma^{s}}\Sigma^{t}\sqrt{\Sigma^{s}})^{1/2}],		
		\end{aligned}    
		\label{cf}
	\end{equation}
	where $\mathrm{Tr}(\cdot)$ is the trace of the matrix. Compared to Euclidean distance, Eq. (\ref{cf}) better reflects the complex distributional discrepancy in the hidden space between the source and target graphs. Thus, the optimal objective of reducing the inter-graph empirical Wasserstein distance corresponds to
	\begin{equation}
		\mathcal{L}_{wass}=\mathcal{D}_{w}(\hat{X}^{s},\hat{X}^{t}).
		\label{lwass}
	\end{equation}
	\subsection{PNDS Information Reduction}
	
	For generating cross-network domain-invariant graph representation, only the CNDI information is useful. To ensure that $\hat{X}^{s}$ and $\hat{X}^{t}$ contain as much CNDI information as possible, we need to eliminate the private domain information they imply (the second optimization path). As such, we additionally adopt a graph-independent encoder (GIE) to construct another attribute transformation of the graph as an anchor attribute matrix, expecting that such a transformation can purely preserve private-network domain-specific(PNDS) information via self-supervised-learning-based graph encoders. As a classic self-supervised graph learning model, deep graph infomax (DGI) \cite{velivckovic2018deep} is selected as an off-the-shelf GIE. Two independent GIEs generate the private node embedding matrices, and the $i+1$th layer is calculated by:
	\begin{equation}
		\begin{aligned}	
			&\tilde{X}^{s(i+1)} = \text{GIE}\left( A^s,\tilde{X}^{s(i)},\tilde{W}^{s(i)} \right),\\
			&\tilde{X}^{t(i+1)} = \text{GIE}\left( A^t,\tilde{X}^{t(i)},\tilde{W}^{t(i)} \right),\\
		\end{aligned}	
		\label{gie}
	\end{equation}
	where $\tilde{X}^{s(0)}=X^s$ and $\tilde{X}^{t(0)}=X^t$. To minimize the mutual information between the two independent graph attribute transformations, we propose the technique of PNDS information reduction.
	
	\begin{myDef}
		\textbf{PNDS information reduction}. If $\hat{X}$ and $\tilde{X}$ are orthogonal in a unified vector space, we say the PNDS information of $\hat{X}$ is completely reduced. For the source and target graphs, we have\\
		\begin{equation}
			\begin{aligned}
				\hat{X}^{s*}= \{\hat{X}^{s}|\hat{X}^{s} \perp \tilde{X}^{s}\},\\
				\hat{X}^{t*}= \{\hat{X}^{t}|\hat{X}^{t} \perp \tilde{X}^{t}\}.
			\end{aligned}
		\end{equation}
	\end{myDef}
	
	To implement PNDS information reduction, we further introduce the Orthogonal Isolation (OI) constraint:
	\begin{equation}
		\begin{aligned}
			\mathcal{L}_{oi} &= l^{s}_{oi}+l^{t}_{oi} \\
			&= \left\| (\hat{X}^{s})^\mathrm{T}\Pi^{s} \tilde{X}^{s} \right\|_{F}^{2} + \left\| (\hat{X}^{t})^\mathrm{T}\Pi^{t} \tilde{X}^{t} \right\|_{F}^{2},
		\end{aligned}
	\end{equation}
	where $\Pi^{s}$ and $\Pi^{t}$ are learnable subspace transformation matrices to align the space coordinates. In the OI constraint, private embedding matrices $\tilde{X}^{s}$ and $\tilde{X}^{t}$ act as the contrastive domains. The minimization of the constraint will result in better graph domain-invariant representations by distancing $\hat{X}^{s}$ and $\hat{X}^{t}$ from the anchor attribute matrices $\tilde{X}^{s}$ and $\tilde{X}^{t}$ such that the PNDS information can be eliminated.
	

	\subsection{Label Space Alignment}
	For the third optimization path, due to the node labels in the target graph being unavailable, the domain alignment in the label space is not trivial. By fully utilizing the existing label information in the source graph, we adopt two prediction tasks for label-related graph domain adaptation, including classification loss $l_{cls}$ in the source graph and entropy loss $l_{entropy}$ in the target graph \cite{wu2020unsupervised}. 
	
	The two tasks can be combined as a label-related optimal objective
	\begin{equation}\mathcal{L}_{label}=l_{cls}+\eta l_{entropy}.
		\label{loss_task}
	\end{equation}
	Note that if the total number of training epochs is $w$ and the current training step is $v\le w$, to ensure stable training, we specially set a self-adaptive parameter $\eta=v/w$ for the second term $l_{entropy}$ in  Eq. (\ref{loss_task}). 
	
	\subsection{Optimizations for GCEs and GIEs}
	For the GCEs, our model aims to jointly minimize the following objectives to obtain the optimal CNDI representations for the source and target graphs.
	\begin{equation}
		{\arg\min}_{\Theta_{oi},\Theta_{label},\Theta_{wass}}\{\mathcal{L}_{oi}+\mathcal{L}_{label}+\lambda \mathcal{L}_{wass}\}
		\label{joint_obj},
	\end{equation}
	where $\lambda$ is a trade-off hyper-parameter that balances the effects of different loss terms. 
	
	Meanwhile, the GIEs can be optimized by minimizing the standard DGI losses, marked as $l_{dgi}^{s}$ and $l_{dgi}^{t}$ for source and target graphs. Note that the GIEs must be trained before the GCEs because the obtained PNDS representations $\hat{X}^{s}$ and $\hat{X}^{t}$ will be harnessed in $\mathcal{L}_{oi}$ as the contrastive domains.
	
	\subsection{Algorithm Description}
	
	The detailed description of the proposed SATMC is presented in Algorithm 1.
	\renewcommand{\algorithmicrequire}{\textbf{Input:}}
	\renewcommand{\algorithmicensure}{\textbf{Output:}} 
	\begin{algorithm}
		\small
		\caption{Structural-attribute transformation with Markov-chain.}
		\label{alg:A}
		\begin{algorithmic}
			\Require ~~\\
			$X^s, X^t$: attribute matrices of the source and target graphs;\\
			$A^s, A^t$: adjacency matrices of the source and target graphs;\\
			$\lambda$: hyper-parameter;
			\Ensure ~~\\
			$\hat{X^s}, \hat{X^t}$: graph representation matrices;
		\end{algorithmic}
		\begin{algorithmic}[1]
			\State {using Eq. (\ref{gie}) to generate PNDS information preserved graph attribute matrices $\tilde{X}^s, \tilde{X}^t$ via self-supervised learning;}\textcolor{blue}{\Comment{GIE encoding}}
			\While {formula (\ref{joint_obj}) not converged}
			\State {using Eq. (\ref{gce}) to generate new graph attribute matrices $\hat{X}^s, \hat{X}^t$;}\textcolor{blue}{\Comment{GCE encoding}}
			\State {calculating $\mathcal{L}_{oi}$ to perform PNDS information reduction;}
			\State {calculating $\mathcal{L}_{wass}$ to reduce empirical Wasserstein distance;}
			\State {calculating $\mathcal{L}_{label}$ to align label space;}
			\State {calculating gradients $\frac{\partial \mathcal{L}_{oi}}{\partial \Theta_{oi}}$, $\frac{\partial \mathcal{L}_{wass}}{\partial \Theta_{wass}}$, $\frac{\partial \mathcal{L}_{label}}{\partial \Theta_{label}}$ and backward propagation via SGD;}
			\EndWhile\\
			\Return{$\hat{X^s}, \hat{X^t}$.}
		\end{algorithmic}
	\end{algorithm}
	
	\subsection{Model Complexity Analysis}
	The proposed model using the PPR kernel in the graph diffusion phase involves calculating the pseudo-inverse of a sparse matrix. Given the sparse matrix has a low rank $M\ll N$ where $N$ is the number of nodes, the pseudo-inverse mainly calculates low-rank singular value decomposition (SVD) \cite{jung2020fast}, which has a complexity $\mathcal{O}(M^2N)$. The encoding side includes GCE and GIE, both involving graph convolution operations, and the time cost of this part is $\mathcal{O}(\left|E\right|d)$ where $d$ is the dimension of the node embeddings. The $\mathcal{L}_{task}$ and $\mathcal{L}_{wass}$ in the formula (\ref{joint_obj}) are mainly responsible for the calculation of cross-entropy (or transformed cross-entropy), and the time cost is at the scale of $\mathcal{O}\left(Nc\right)$, where $c$ is the number of categories. Due to the source and target graphs having a similar process, the total complexity is $\mathcal{O}(\max({(M^{s})}^2N^{s}, (M^{t})^2N^{t})+\max(\left|E^{s}\right|,\left|E^{t}\right|)d+\max(N^s,N^t)c)$, which is much better than the quadratic complexity of the previous GDA methods \cite{shi2023improving,zhang2021adversarial}.

	\subsection{Role of PPR Diffusion Kernel and Wasserstein Graph Distance}
	Theoretically, we give an in-depth study on why introducing the PPR-kernel-based graph diffusion and reducing the Wasserstein distance between graph representations are beneficial to the cross-network node classification task.
	
	\newtheorem{p1}[theorem]{Lemma}
	\begin{p1}
		If the graph Laplacian has the spectral radius $\rho(L)<1$, the graph diffusion matrix with PPR kernel is $\alpha$-Lipschitz continuous w.r.t. the graph Laplacian.
		
	\end{p1}
	\begin{proof}
		By applying the PPR kernel with teleport probability $\alpha \in (0,1)$\cite{chung2007heat}, the graph diffusion mapping has the general form
		\begin{equation}
			\hat{A} = {\sum\limits_{k = 0}^{\infty}\,}\alpha(1 - \alpha)^{k}\mathcal{T}^{k}.
		\end{equation}
		where the symmetric transition matrix meets $\mathcal{T} = D^{- 1/2}AD^{- 1/2}$ and $D$ is the diagonal degree matrix. Based on the spectral analysis in \cite{gasteiger2019diffusion}, we have an equivalent form
		\begin{equation}
			{\sum\limits_{k = 0}^{\infty}\,}\alpha(1 - \alpha)^{k}\mathcal{T}^{k}={\sum\limits_{j = 0}^{\infty}}(1 - \frac{1}{\alpha})^{j}L^{j}.
		\end{equation}
		For the graph diffusion matrix with different perturbations, it meets that 
		\begin{equation}
			\begin{aligned}
				\left\|\hat{A}'-\hat{A}''\right\|_{F}&=\left\|{\sum\limits_{j = 0}^{\infty}\,}(1 - \frac{1}{\alpha})^{j}(L')^{j}-{\sum\limits_{j = 0}^{\infty}\,}(1 - \frac{1}{\alpha})^{j}(L'')^{j}\right\|_{F} \\
				&=\left\|{\sum\limits_{j = 0}^{\infty}\,}(1 - \frac{1}{\alpha})^{j}(\underbrace{(L')^{j}-(L'')^{j} }_{\rho(L)<1})\right\|_{F} \\
				& \le \left\|{\sum\limits_{j = 0}^{\infty}\,}(1 - \frac{1}{\alpha})^{j}(L'-L'')\right\|_{F}. \\
			\end{aligned}
		\end{equation}
		Recognize the series expansion $\frac{1}{1-x}=\sum_{i=0}^{\infty}x^{i}$, resulting in
		\begin{equation}
			\left\|\hat{A}'-\hat{A}''\right\|_{F} \le \alpha \left\|L'-L''\right\|_{F}.
		\end{equation}	
		Therefore, $\hat{A}$ is $\alpha$-Lipschitz continuous w.r.t. the graph Laplacian.
	\end{proof}
	
	Let a hypothesis class $H$ be a set of bounded real-value functions denoting different node classifiers, and each $h\in H$ satisfies $h: \hat{X}\to Y$ given a graph projection function $\phi: G\to\hat{X}$. The classification error can be defined as $\epsilon(h)=\mathbb{E}_{x\in \hat{X}} [\mathcal{L}(h(x),y_{x})] $ where $y_{x}\in \mathbb{C}$ is the class of $x$ and $\mathcal{L}(\cdot)$ is the classification loss.
	\newtheorem{p2}[theorem]{Theorem}
	\begin{p2}
		Suppose the graph projection functions $\phi_s$ and $\phi_t$ are $\alpha$-Lipschitz continuous w.r.t. the graph Laplacians. For a node classifier hypothesis class $H$, $\forall h\in H$, it has a generalized error bound as
		\begin{equation}
			\epsilon_{t}(h) \le \epsilon_{s}(h)+2\alpha\mathcal{L}_{wass}+\kappa ,
		\end{equation}
		where $\kappa $ is the combined error of the ideal hypothesis $h^{*}$.
		
	\end{p2}
	\begin{proof}
		According to the Kantorovich-Rubinstein Theorem \cite{villani2009optimal}, the first Wasserstein distance of classification error distributions between the source graph and target graph can be defined as
		\begin{equation}
			\mathcal{D}_{w}(\epsilon_{s}(h),\epsilon_{t}(h))= \sup_{\|h\|_{Lip} \le 1} |\mathbb{E}_{s} (h \circ \phi_s)-\mathbb{E}_{t} (h \circ \phi_t)|.
		\end{equation}
		With the observation that $h \circ \phi_s$ and $h \circ \phi_t$ are the extensions of $\phi_s$ and $\phi_t$, which are also $\alpha$-Lipschitz continuous w.r.t. the graph Laplacians on the basis of Kirszbraun Theorem \cite{valentine1945lipschitz}. Since the distance of the embedding distributions between source and target graphs is measured by the empirical Wasserstein distance, according to Theorem 1 shown in \cite{shen2018wasserstein} and Lemma 1, we derive that
		\begin{equation}
			\begin{aligned}
				\epsilon_{t}(h) &\le \epsilon_{s}(h)+2\alpha \mathcal{D}_{w}(\hat{X}^{s},\hat{X}^{t})+\kappa  \\
				&=\epsilon_{s}(h)+2\alpha\mathcal{L}_{wass}+\kappa .
			\end{aligned}
		\end{equation}
	\end{proof}
	Theorem 2 indicates that the classification error in the target graph can be reduced by minimizing both the classification error in the source graph and the Wasserstein distance between the source graph domain and the target graph domain. As $\alpha \in (0,1)$, the PPR-kernel-based graph diffusion can ensure a tighter error bound than traditional graph domain adaptation methods like GRADE-N \cite{wu2023non}.
	
	\section{Experiments}
	To assess the effectiveness of the proposed framework, we have performed comprehensive experiments to answer five research questions.\\
	RQ1: Can the proposed new model outperform the existing state-of-the-art graph domain adaptation methods in the cross-network node classification task?\\
	RQ2: Which component of the proposed framework is more significant? \\
	RQ3: Does the model converge faster than the existing methods?\\
	RQ4: Is PNDS information reduction effective for constructing domain-invariant features between the source and target graphs?\\
	RQ5: Can Wasserstein graph domain adaptation eliminate the feature discrepancies  between nodes of the same category in the source and target graphs?\\

	\subsection{Datasets}
	
	To evaluate node classification across networks, we use six benchmark datasets: four citation networks including Citationv1, DBLPv7, ACMv9 \cite{tang2008arnetminer}, and Arxiv \cite{liu2023structural}, plus two social graphs including Blog1 and Blog2 (B1, B2 for short) from BlogCatalog \cite{li2015unsupervised}.
	
	In the citation networks (except Arxiv), nodes are papers, edges are citations, attributes are keywords from titles, and labels denote research fields. We consider six cross-network tasks: A$\to$C, A$\to$D, C$\to$A, C$\to$D, D$\to$A, and D$\to$C, where, e.g., “A$\to$C” means ACMv9 as source and Citationv1 as target. Arxiv contains CS papers from 40 categories, with attributes as embeddings of titles and abstracts. Following Liu et al. \cite{liu2023structural}, papers from 2011–2014 form the source and 2018–2020 the target, enabling temporal domain-shift evaluation.
	
	For B1 and B2, nodes are bloggers, edges are friendships, attributes are self-description keywords, and labels mark group membership. Since both networks share distributions, domain discrepancy is introduced by flipping 30\% of non-zero attributes to 0 and 30\% of zeros to 1, as in Shen et al. \cite{shen2020adversarial}, simulating incomplete or noisy attributes. Dataset statistics are summarized in Table \ref{table2}.

	\begin{table}[h]
		\centering
		\caption{Statistical information of the six datasets.}
		
		{
			\begin{tabular}{c|ccccc}
				\toprule
				\textbf{Datasets}   & \textbf{\#Nodes} & \textbf{\#Edges} & \textbf{\#Attributes}  & \textbf{\#Labels} \\ \hline
				DBLPv7     & 5,484   & 8,130   & 4,412                      & 5       \\
				ACMv9      & 9,360   & 15,602  & 5,571                     & 5       \\
				Citationv1 & 8,935   & 15,113  & 5,379                       & 5       \\
				Blog1      & 2,300   & 33,471  & 8,189                       & 6       \\
				Blog2      & 2,896   & 53,836  & 8,189                       & 6       \\ 
				Arxiv   &169,343 & 2,315,598  & 128                      & 40     \\
				\bottomrule
			\end{tabular}
		}
		\label{table2}
	\end{table}

	\subsection{Baselines}
	\textbf{Source-only approaches}. GCN\cite{kipf2016semi} and GSAGE\cite{hamilton2017inductive} use only the source graph labeling data, which are trained on the source graph and tested directly on the target graph.\\
	\textbf{Conventional domain adaptation approaches}. DANN\cite{ganin2016domain} is a representative method in traditional domain adaptation modeling, which does not consider the intrinsic relationships between samples.\\
	\textbf{Graph domain adaptation approaches}. The GNN-based GDA models proposed in recent years include AdaGCN\cite{dai2022graph}, UDAGCN\cite{wu2020unsupervised}, ASN\cite{zhang2021adversarial}, GRADE-N\cite{wu2023non}, SA-GDA\cite{pang2023sa}, SGDA\cite{qiao2023semi}, CMPGNN\cite{wang2025bridging}, SpecReg\cite{you2022graph}, StruRW-ERM \cite{liu2023structural} and JHGDA\cite{shi2023improving}.
	
	\subsection{Experimental settings}
	
	Training epochs are adjusted per dataset. GCE uses the PPR kernel with teleport probability $\alpha=0.05$, and constructs a high-order adjacency matrix via threshold truncation ($\xi$=1e-3). The joint loss parameter $\lambda$ is dataset-specific. Model adopts Adam optimization with the learning rate 0.02 for citation networks and 0.001 for social graphs, and the weight decay is 5e-4. All models use 2 graph convolutional layers. For citation datasets, hidden and output dimensions are 128 and 16; for Blog networks, 512 and 256. The dropout ratio is fixed at 0.5. Results are reported by accuracy (ACC), averaged over 5 runs. Detailed parameters appear in Table \ref{table_params}.

	\begin{table}[h]
		\centering
		\caption{Detailed parameter settings for the experiments.}
		\begin{tabular}{c|ccccccc}
			\toprule                                                    
			\textbf{Dataset pair}     & \textbf{Epoch}             & \textbf{$\lambda$}             & \textbf{Learning rate}            \\ \hline
			A$\to$D     & 100           & 0.2           & 0.02          \\
			D$\to$A   & 100           & 0.1           & 0.02      \\
			A$\to$C    & 100           & 0.1           & 0.02       \\
			C$\to$A  & 100           & 0.1           & 0.02  \\
			C$\to$D  & 250           & 0.1           & 0.02    \\
			D$\to$C    & 200           & 0.1          & 0.02      \\
			B1$\to$B2 & 500           & 0.1            & 0.001  \\
			B2$\to$B1 & 500          & 0.15            & 0.001    \\
			Arxiv & 1000          & 0.001            & 0.001    \\
			\bottomrule
		\end{tabular}
		\label{table_params}
	\end{table}
	
	For baselines, the numbers of GNN layers and dimensions of hidden/output layers match ours, while other parameters follow the original papers. Each method, including ours, trains a single-layer MLP classifier on the source graph to test target graph accuracy. StruRW-ERM follows a semi-supervised setting, using 20\% target labels for validation and 80\% for testing, whereas other baselines adopt unsupervised domain adaptation, training only on the source graph. All experiments ran on an Intel(R) Xeon(R) 2.20GHz CPU with NVIDIA TITAN RTX GPUs (24GB).

	\subsection{Main Results (RQ1)}
	Based on the results of the comparative experiments shown in Table \ref{main_results}, we mainly report the following observations:
	
	(1) By merely using attribute features without considering the topological information of the graph, the DANN method obtained an averaged accuracy $49\%$ over citation networks and $41.5\%$ over social networks, which has the worst performance for across-network node classification.
	
	\begin{table*}[h]
		\centering
		\caption{Accuracy comparison of cross-network node classification on nine pairs of cross-domain datasets. The best and second-best results are marked in boldface and underlined, respectively. OOM denotes that the model training is out of memory.}
		\begin{threeparttable}
			\begin{tabular}{c|ccccccc|ccc|c}
				\toprule
				\multicolumn{1}{c|}{\textbf{Methods}} &\multicolumn{7}{c|}{\textbf{Citation Networks}}  & \multicolumn{3}{c|}{\textbf{Social Networks}}  & \multicolumn{1}{c}{\textbf{Large Network}}                                                                                                         \\ \hline
				& A$\to$D             & D$\to$A             & A$\to$C             & C$\to$A            & C$\to$D             & D$\to$C             & Avg.            & B1$\to$B2             & B2$\to$B1             & Avg.     & Arxiv      \\
				GCN      & 0.623           & 0.578           & 0.675           & 0.635          & 0.666           & 0.654          & 0.638    & 0.501           & 0.467           & 0.484    &0.579  \\
				GSAGE   & 0.665        & 0.593           & 0.717           & 0.653          & 0.701           & 0.685           & 0.669   & 0.46           & 0.446           & 0.453   &0.592   \\
				DANN    & 0.488           & 0.436           & 0.52            & 0.518          & 0.518           & 0.465            & 0.49    & 0.41           & 0.419           & 0.415   &0.488   \\
				AdaGCN  & 0.687           & 0.663           & 0.701           & 0.643          & 0.709           & 0.702            & 0.684   & 0.522           & \underline{0.532}           & 0.527    &0.580  \\
				UDAGCN  & 0.684           & 0.623           & 0.728           & 0.663          & 0.712           & 0.645            & 0.675     & 0.522           & 0.517           & 0.519  &0.561   \\
				ASN     & 0.729           & 0.723           & 0.752           & 0.678          & 0.752           & 0.754           & 0.731    & 0.515           & 0.498           & 0.506   & OOM  \\
				GRADE-N & 0.701           & 0.66            & 0.736           & 0.687          & 0.722           & 0.687            & 0.698    & 0.507           & 0.473            & 0.49   &0.488   \\
				SA-GDA & 0.644           & 0.63            & 0.642           & 0.614          & 0.71           & 0.58            & 0.637    & 0.19           & 0.174            & 0.182   &OOM   \\
				SGDA & 0.701           & 0.642            & 0.779           & 0.659          & 0.726           & 0.784            & 0.715    & 0.243           & 0.226            & 0.234   &OOM   \\
				CMPGNN & 0.695           & 0.597            & 0.744           & 0.639          & 0.722           & 0.722            & 0.686    & 0.174           & 0.182            & 0.178   &OOM   \\
				SpecReg & 0.662           & 0.55            & 0.733           & 0.582          & 0.668           & 0.697            & 0.648     & 0.383           & 0.376            & 0.379  &\underline{0.602}   \\
				StruRW-ERM & 0.5627 & 0.515 & 0.575 & 0.55  & 0.605 & 0.585   & 0.565  & 0.349  & 0.414  & 0.382  & 0.489 \\
				JHGDA   & \underline{0.754}           & \underline{0.737}           & \underline{0.814}           & \textbf{0.757} & \underline{0.762}           & \underline{0.794}           & \underline{0.769}      & \underline{0.544}           & 0.53           & \underline{0.537}     & OOM \\ \hline
				SATMC\footnotemark[1]    & \textbf{0.779} & \textbf{0.741} & \textbf{0.828} & \underline{0.744}         & \textbf{0.779} & \textbf{0.81}   & \textbf{0.78 }  & \textbf{0.551} & \textbf{0.589} & \textbf{0.57} &\textbf{0.612 } \\ 
				\bottomrule
			\end{tabular}
			\centering
			\begin{tablenotes}            
				\item[1] The paired T-test with $p=0.015$ indicates that SATMC outperforms the best baseline (JHGDA) with statistical significance at the 95\% confidence interval.
			\end{tablenotes}
		\end{threeparttable}
		\label{main_results}

	\end{table*}
	
	(2) The results obtained by GCN and GSAGE are also unsatisfactory, which roughly have an average $10\%$ margin behind SATMC on both citation networks and social networks. This indicates that the distribution discrepancy between the source and target graphs significantly affects the classification performance of the classifier on the target domain. Consequently, it is insufficient to simply apply classifiers trained on the source graph to classify nodes in the target graph.
	
	(3) All GDA methods outperform the source-only and traditional domain adaptation approaches, indicating that simultaneous use of source and target graph information for cross-network node classification tasks allows for better domain adaptation. However, we also notice the significant differences between our method and others. For instance, despite using Graph Subtree Discrepancy (GSD) to measure the domain divergence between the source and target graph \cite{wu2023non}, GRADE-N has the poorest performance among graph domain adaptation methods, indicating that the domain divergence is not sufficiently reduced. Through both structure and attribute alignments, our method achieves the best results on eight pairs of datasets and improves the average margin by 2.4\% against the SOTA baseline JHGDA. In particular, the improvements exceed by 3.1\% on A$\to$D and up to 6.2\% on B2$\to$B1, respectively.
	
	(4) For the large network Arxiv, four baselines encountered the problem of memory exhaustion, among which JHGDA was the best-performing method for citation networks and social networks. This suggests that our method not only has strong domain adaptability but also has fair scalability.

	\subsection{Ablation Study (RQ2)}

	\begin{table}[h]
		\scriptsize
		\centering
		\caption{Ablation experiments on six pairs of citation networks.}
		\begin{tabular}{cccccc}
			\toprule
			& w/o NGDC & w/o $\mathcal{L}_{wass}$ & w/o $\mathcal{L}_{label}$ & w/o $\mathcal{L}_{oi}$ & SATMC \\
			\hline
			A $\to$ D & 0.623 & 0.734  & 0.745 & 0.769  & \textbf{0.779 } \\
			D $\to$ A & 0.578 & 0.700  & 0.707 & 0.721  & \textbf{0.741 } \\
			A $\to$ C & 0.675 & 0.796  & 0.81  & 0.811  & \textbf{0.828 } \\
			C $\to$ A & 0.635 & 0.735  & 0.75  & 0.741  & \textbf{0.744 } \\
			C $\to$ D & 0.666 & 0.750  & 0.778 & 0.772  & \textbf{0.779 } \\
			D $\to$ C & 0.654 & 0.757  & 0.795 & 0.808  & \textbf{0.81 } \\
			\hline
			Avg.  & 0.639  & 0.744  & 0.764  & 0.77  & \textbf{0.78 } \\
			\bottomrule
		\end{tabular}
		\label{table_ablation}
	\end{table}
	To understand the role of each part of the proposed GDA framework, we performed ablation experiments on the citation datasets. We first investigate the items in the joint objective formula (\ref{joint_obj}). According to Table \ref{table_ablation}, if removing the item of PNDS information reduction $\mathcal{L}_{oi}$, the model's node classification accuracy on the target network decreases by 1\% on average. When further removing the label-related loss item $\mathcal{L}_{label}$, the model's accuracy shows a slightly drop by 0.6\% on average. If the Wasserstein distance term $\mathcal{L}_{wass}$ is again removed, the accuracy of the model drops by an average of 2\%. If further removing the structure-attribute transformation module, i.e., the NGDC, the model will degenerate to the vanilla GCN, and the accuracy significantly drops by $10.5\%$ on average. The ablation experiments suggest that the NGDC plays the most important role in our framework. For model optimization, the terms $\mathcal{L}_{wass}$ and $\mathcal{L}_{oi}$ in formula (\ref{joint_obj}) contribute more significantly than the $\mathcal{L}_{label}$.

	\subsection{Convergence of Model Performance (RQ3)}
	\begin{figure}[h]
		\centering
		\includegraphics[width=1\linewidth]{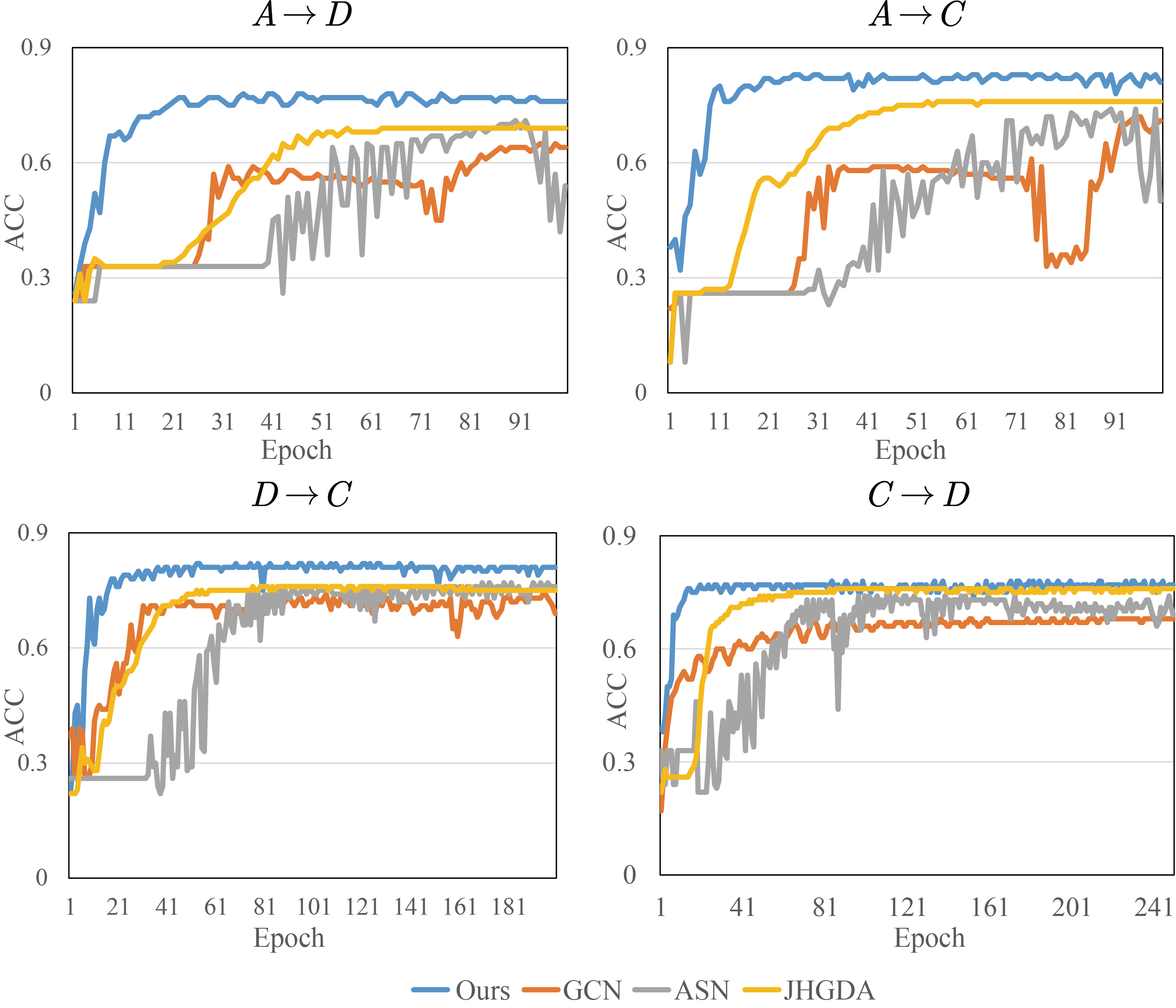}
		\caption{ACC curves along with the training steps on four pairs of citation networks.}
		\label{loss}
	\end{figure}
	
	In order to analyze the convergence of model performance, w.o.l.g., we select four pairs of datasets, A$\to$D, A$\to$C, D$\to$C and C$\to$D, to demonstrate the accuracy curves with limited training epochs between our method and three baselines (100 epochs on A$\to$D and A$\to$C while 200 epochs on D$\to$C and 250 epochs on C$\to$D). It is shown in Fig. \ref{loss} that, both our method and JHDGA can converge to steady ACCs while the ACC curves for GCN and ASN are quite unstable throughout the training process. Compared to JHDGA, our method can reach the optimal ACCs with fewer than $25\%$ of training epochs, indicating it has higher training efficiency. For instance, on the A$\to$D, only 11 epochs are required for our method to converge whereas 50 epochs for JHDGA. For A$\to$C, D$\to$C and C$\to$D, we can observe the similar results. From Fig. \ref{loss}, we also notice that for the four groups of datasets, our method has consistently better ACCs against the three other baselines during the training.

	\subsection{Validation of Domain Invariant Representation (RQ4)}
\begin{figure}[h]
		\centering
		\includegraphics[width=1\linewidth]{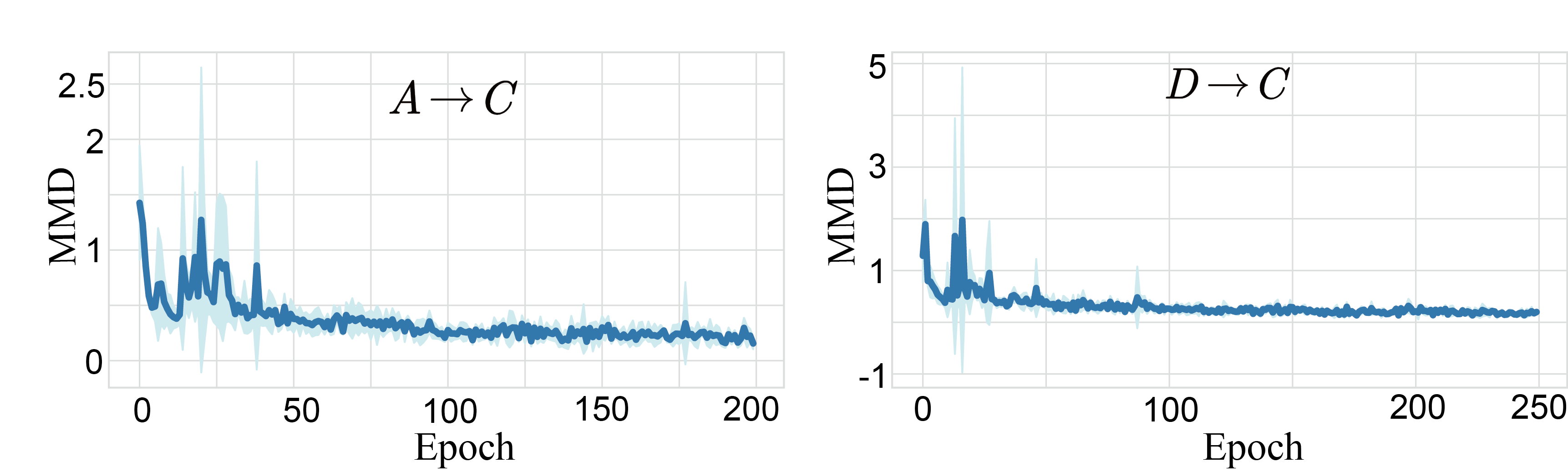}
		\caption{MMD curve solely relying on PNDS. The shaded area represents the standard error.}
		\label{mmd}
	\end{figure}

	The Maximum Mean Discrepancy (MMD) is commonly used to assess the domain invariance between source and target domain distributions \cite{borgwardt2006integrating,arbel2019maximum}. A smaller MMD value indicates that the source and target domains share more domain-invariant information. To observe the evolutionary trend of MMD during GCE training, we disable other modules and only retain the PNDS information elimination module. As shown in Fig. \ref{mmd}, on both datasets A$\to$C and D$\to$C, the MMD oscillates in the initial 50 epochs and generally converges in the remaining training epochs with standard error dropped by 92\% on A$\to$C and 98\% on D$\to$C averagely. The MMD curves suggest that, despite the gradual reduction of PNDS information, the fluctuations of MMD values imply the complexity of reducing domain inconsistency.
	
	\subsection{Improvement of Category Discrepancy (RQ5)}
\begin{figure}[h]
		\centering
		\includegraphics[width=1\linewidth]{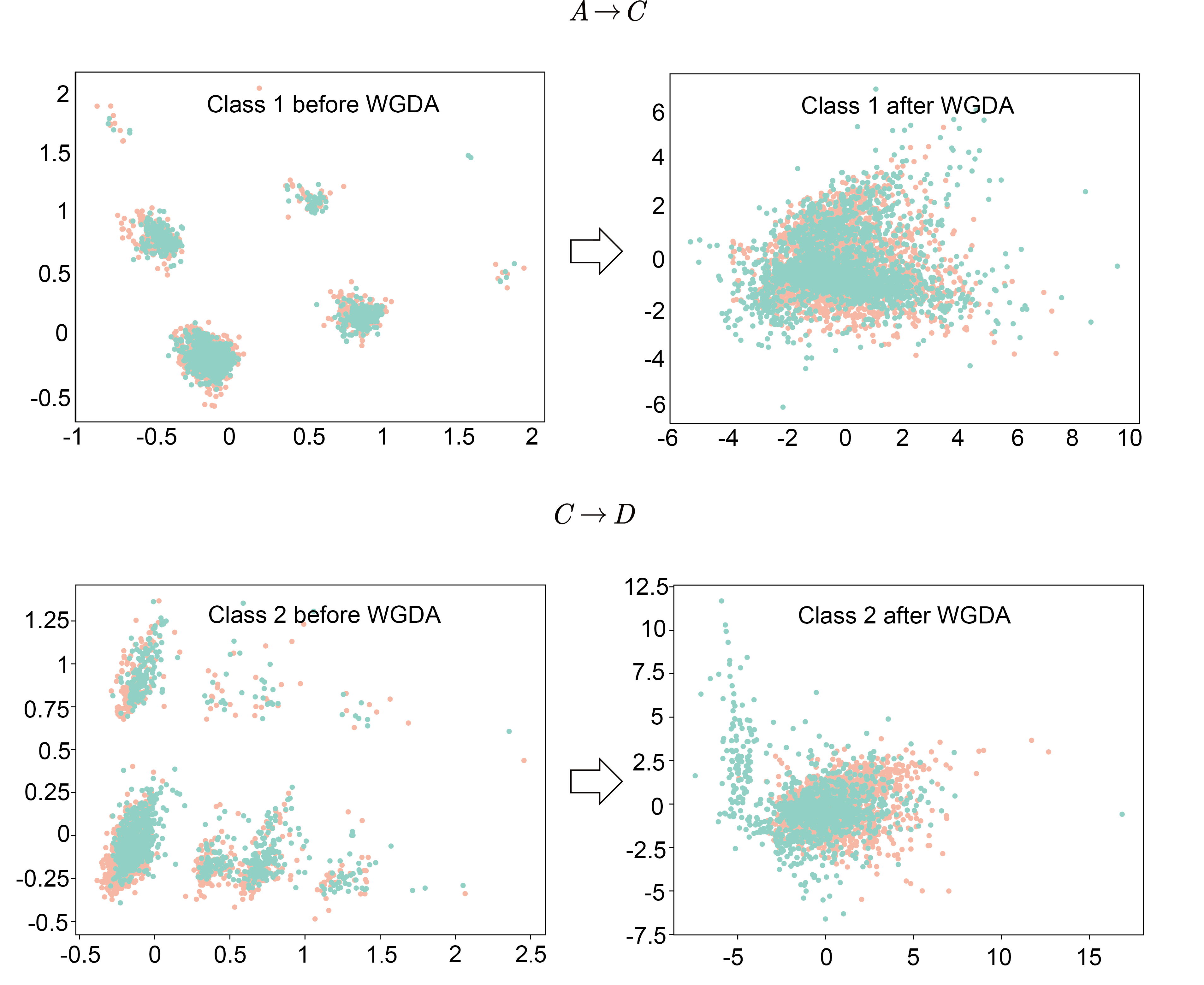}
		\caption{Comparison of single-category scatter distribution before and after WDGA.}
		\label{single_class}
	\end{figure}

	For analytical convenience, we select one-class data points from each of the datasets A$\to$C and C$\to$D to examine their distribution shifts before and after applying WGDA. As illustrated in Fig. \ref{single_class}, prior to WGDA, the data points of class 1 are scattered across 4 clusters. This separation could lead to class-1 nodes being misclassified as 4 distinct classes. In contrast, after applying WGDA, the points merge into a single and cohesive cluster, indicating closer spatial proximity and thereby reducing the likelihood of misclassification. Similarly, class-2 data points in C$\to$D display analogous changes in distribution after WGDA. This visualization confirms that our method can effectively resolve research question 5.
	
	\subsection{Visualization}
	\begin{figure}[t]
		\centering
		\includegraphics[width=1\linewidth]{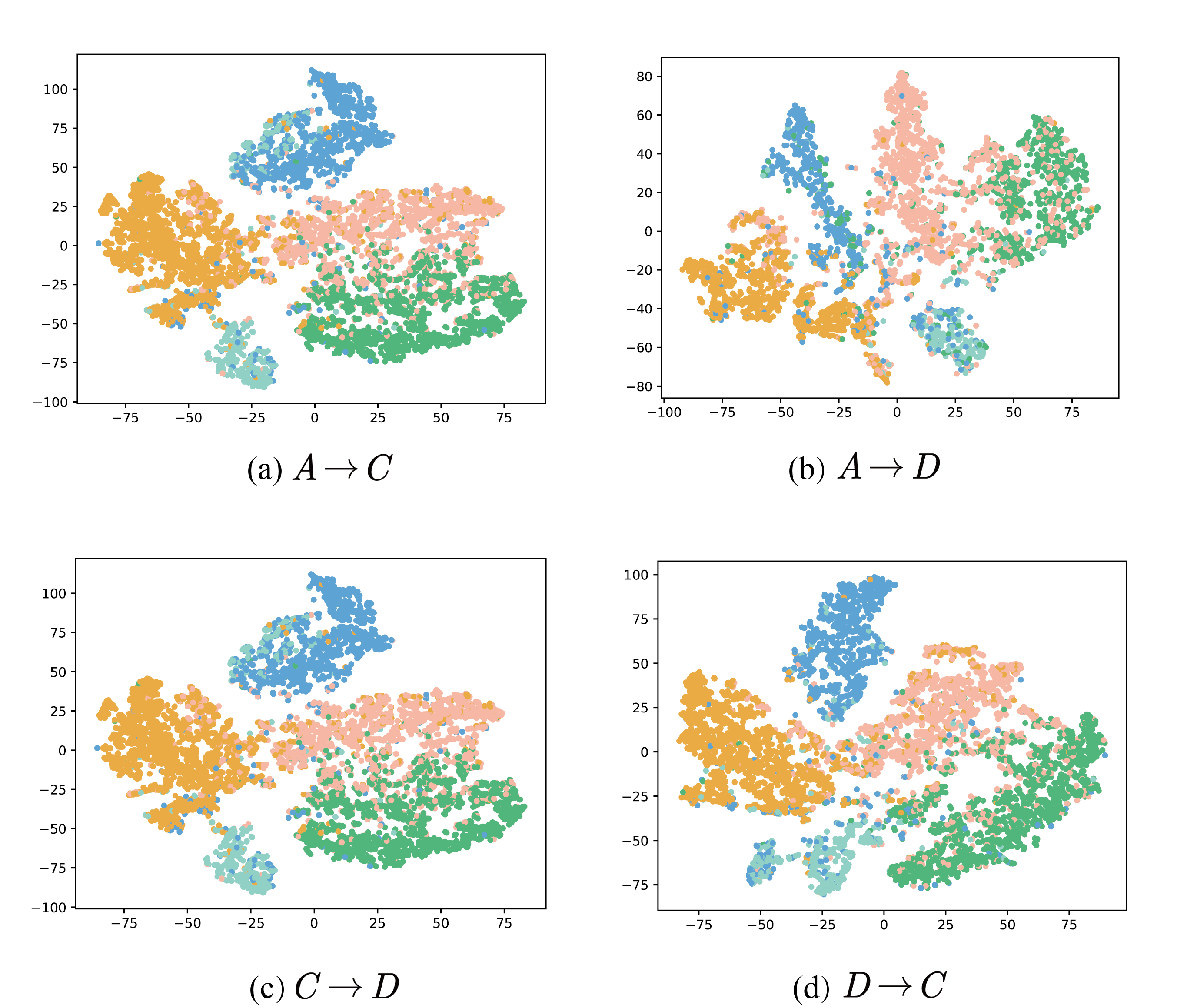}
		\caption{Visualization of the 2D target embedding distributions on four pairs of citation networks.}
		\label{visualization}
	\end{figure}
	To visualize the domain adaptation efficacy, we use t-SNE \cite{van2008visualizing} to project the target node embedding vectors into a 2D space. For the four pairs of citation datasets, as illustrated in Fig. \ref{visualization}, the points with the same color are clustered together, demonstrating good category separability. This indicates that the domain adaptations between the source and target graphs were successful, as the node label information from the source graph has been effectively transferred to the unlabeled nodes of the target graph.

	\section{Conclusion}
	This paper introduces SATMC, a novel graph transfer learning framework tailored for unsupervised graph domain adaptation (GDA). In contrast to conventional GDA approaches that primarily emphasize attribute consistency, SATMC simultaneously addresses both structural and attribute alignments. By employing a Markov chain-based structure-attribute graph transformation, the framework achieves a tighter theoretical error bound for the cross-network node classification task. Extensive experiments on 9 pairs of cross-domain datasets demonstrate that SATMC outperforms existing state-of-the-art GDA methods in terms of domain adaptation performance. Furthermore, ablation studies highlight the critical role of graph structure transformation within the framework, underscoring the importance of aligning topological distributions in effective graph domain adaptation.
	
	\begin{acks}
		This work is supported by the SiChuan Provincial Natural Science Foundation (No. 2024NSFSC0517).
	\end{acks}
	
	\section*{GenAI Usage Disclosure}
	We declare that AI tools were used only to enhance phrasing and readability, while all research methods, data analyses, and findings are entirely the authors’ own contributions.
	
	\bibliographystyle{ACM-Reference-Format}
	\balance
	\bibliography{reference}

\end{document}